\documentclass[11pt]{article}

\usepackage[final]{acl}

\usepackage{times}
\usepackage{latexsym}

\usepackage{tabularx}

\usepackage[T1]{fontenc}

\usepackage[utf8]{inputenc}

\usepackage{microtype}

\usepackage{inconsolata}

\usepackage{graphicx}

\usepackage[nolist]{acronym}

\usepackage{enumitem}
\setlist[itemize]{noitemsep, nosep}

\DeclareGraphicsExtensions{.pdf,.ai,.jpg,.png}
\setkeys{Gin}{pagebox=artbox}
\graphicspath{{sections/img/}}

\RequirePackage{type1cm}
\RequirePackage{color}
\RequirePackage{soul}
\setstcolor{blue}
\definecolor{violet}{rgb}{0.5,0.0,0.5}
\newsavebox\hwcombox
\newcommand{\hw}[3][]{%
	\sbox{\hwcombox}{\fontsize{8}{9}\selectfont#1#2#3}
	\noindent
	\st{#2}{\selectfont
		\color{blue}#3\ifx\\#1\\\else{\fontsize{8}{9}\selectfont\color{violet}[#1]}\fi
	}
}

\title{Toward Reasonable Parrots: \\ Why Large Language Models Should Argue with Us by Design}

\author{
Elena Musi$^{1}$, 
Nadin Kokciyan$^{2}$, 
Khalid Al-Khatib$^{3}$, 
Davide Ceolin$^4$, \\
\textbf{Emmanuelle Dietz$^5$, 
Klara M. Gutekunst$^6$, 
Annette Hautli-Janisz$^7$}, \\
\textbf{Cristian Manuel Santibáñez$^8$, 
Jodi Schneider$^9$, 
Jonas Scholz$^3$,} \\
\textbf{Cor Steging$^3$,
Jacky Visser$^{10}$,
Henning Wachsmuth$^{11}$}\\ \\[-1.25ex]
$^1$University of Liverpool, 
$^2$University of Edinburgh, 
$^3$University of Groningen, \\
$^4$Centrum Wiskunde \& Informatica,
$^5$Airbus Central R\&T, Hamburg $^6$University of Kassel,\\ 
$^7$University of Passau,
$^8$Universidad de Católica de la Santísima de Concepción, \\
$^9$University of Illinois at Urbana-Champaign, 
$^{10}$University of Dundee, \\
$^{11}$Leibniz University Hannover\\ \\[-1.25ex]
Corresponding author: \texttt{elena.musi@liverpool.ac.uk}\\ \\
}

\begin{document}
\maketitle

\begin{abstract}
In this position paper, we advocate for the development of conversational technology that is inherently designed to support and facilitate argumentative 
processes. We argue that, at present, large language models (LLMs) are inadequate for this purpose, and we propose an ideal technology design aimed at enhancing argumentative skills. This involves re-framing LLMs as tools to exercise our critical thinking skills rather than replacing them. We introduce the concept of \textit{reasonable parrots} that embody the fundamental principles of relevance, responsibility, and freedom, and that interact through argumentative dialogical moves. These principles and moves arise out of millennia of work in argumentation theory and should serve as the starting point for LLM-based technology that incorporates basic principles of argumentation.
\end{abstract}

\maketitle
\section{Introduction}

Currently, \acp{llm} are revolutionizing how we access information, providing instant answers to queries in various helpful ways thanks to their impressive rhetorical capabilities~\cite{hyde2004ethos}. \acp{llm} can mimic a wide range of tones and styles, which can be tailored to user preferences \cite{carrascofarre:2024}. As a result, users may perceive LLM responses as akin to those of trustworthy experts, despite the inherent limitations~\cite{echterhoff2024cognitive, lin2025implicit}. 

LLMs can produce responses that are insufficient, misleading, or downright harmful. From a dialectical perspective, they inherently instantiate the ``ad populum fallacy''~\cite{walton1980ad}: they echo ideas that are sufficiently popular 
in their training data, implicitly suggesting that popularity equates to truth.
This contrasts with the human perspective on critical thinking, where knowledge is contextually grounded and entails a stake in the knowledge produced and its consequences \cite{lindebaum2024chatgpt}.
In this position paper, we advocate for \acp{llm} that \emph{argue} with us by design; conversational technology that can foster deliberation, ask for justifications, avoid fallacies, and give alternatives, in line with theories of reasonableness stemming from argumentation~\cite{van2003pragma,walton:2006}.

\begin{figure}[t]
\centering
\includegraphics[scale=1]{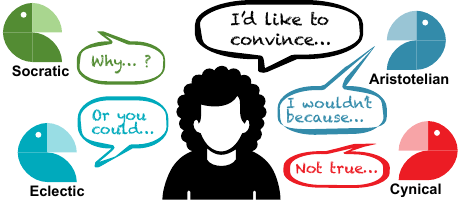}
\caption{Sketch of a possible realization of the idea of reasonable parrots we advocate for: a conversational technology that fosters critical thinking and deliberation.}
\label{parrot-sketch}
\end{figure}

Despite LLMs' impressive capabilities to generate coherent text, \citet{bender:2021} emphasize that LLMs are little more than stochastic parrots: they are programmed to predict the likelihood of a token based on the preceding context, but they do not ``understand'' what they generate. 

As such, the question ``\textit{Can you argue with a parrot?}'' 
sounds rhetorical at first sight, given that parrots are  just repeating what they hear without understanding it. Undoubtedly, though, \acp{llm} are increasingly used in decision-making processes, ranging from medicine \cite{kim2024mdagents} to finance \cite{lakkaraju2023llms} and human resources \cite{sun2024research}. In these scenarios, biased standpoints and fallacious reasoning can have significant societal consequences. Even though \acp{llm} seem able to perform reasoning tasks for which they were not explicitly trained \cite{Emergent2023Webb}, these emerging reasoning capabilities are limited and far from understood \cite{lewis2024evaluating}. The view that LLMs are incapable of understanding is in fact controversial, as it assumes a theory of meaning grounded in reference, which is not essential in language games \cite{li2015word} or in abstract thinking that involves forming internal representations of concepts \cite{piantadosi2022meaning}. However, there is  agreement that, if these systems construct meaning at all, they do so in a manner fundamentally different from humans. Therefore, they should not be anthropomorphized as arguers who share a common ground of knowledge with humans. 

In this paper, we argue that it is essential to move away from \emph{stochastic} parrots toward conversational technologies that interact with humans as \emph{reasonable} parrots~\cite{van2015reasonableness, akata2020research}, as sketched in Figure~\ref{parrot-sketch}. Although they remain parrots, reasonable parrots adopt an attitude of argumentative reasonableness---grounded in the fundamental human principles of relevance, responsibility, and freedom~\cite{danesi2009global}---to help enhance users' critical thinking skills: (i) \textit{Principle of relevance.} Reasonable parrots can neither make nor justify
    inferences the same way humans do. 
    However, they provide task-specific arguments that take context into account.
(ii) \textit{Principle of responsibility.} Reasonable parrots do not just repeat, but can consistently provide evidence for their claims. 
(iii) \textit{Principle of freedom.} Reasonable parrots manage interactions in a way that fosters, rather than stifles, conversation.

In other words, reasonable parrots do not persuade us of a decision, but perform dialogical moves which match the rules of an ideal critical discussion in pragma-dialectics \cite{van2003pragma}, such as expressing doubts, rebutting arguments, or providing alternatives. Thereby, they foster critical thinking and deliberation, challenging our views and steering us toward seeing the full picture. More specifically, they aim to foster both individual critical thinking skills, such as the ability to assess reasons and avoid cognitive biases, and social critical thinking skills, such as reflecting on prevailing social norms \cite{larson2024critical}. In doing so, reasonable parrots challenge both lines of reasoning and commonly accepted knowledge.

\section{State of the Art}


The rhetorical capabilities of \acp{llm} are confirmed by their ability to turn ineffective arguments into effective ones for people with certain ideologies \cite{el2024improving} or to make inappropriate arguments appropriate \citep{ziegenbein:2024a}. \citet{herbold2023large} demonstrated that ChatGPT produces essays that are rated higher in quality than those written by humans. For reasoning, studies show that \acp{llm} do not reason through arguments; rather, they are highly sensitive to levels of abstraction \cite{de2024d} and provide explanations about their reasoning that do not match the process they followed \cite{Steging2021rationale,musi2024fallacy}. 
Neuro-symbolic approaches may improve soundness and explainability, but they are domain-specific and cannot be easily scaled up \cite{ranaldi2025improving}. Additionally, using argumentative frameworks in prompt engineering \cite{freedman2024argumentative} seems to improve general performance, but it does not make \acp{llm} close to humans as discussants. 

A seminal attempt at deliberation before the popularization of \acp{llm} was IBM's Project Debater \cite{slonimAutonomousDebatingSystem2021}, where the authors argue that argumentative debates, unlike games such as chess, fall outside of the comfort zone of AI. 
This is because language in general, and argumentation in particular, are far less structured and rely more on human interpretation, real-world issues, situative setting, and common sense. 
In the context of argument search, \citet{kiesel:2021} suggested personas as an possible way to reduce the selection bias of argumentative conversational technology.

Less attention has been given to the impact that human-\ac{llm} interaction has on human argumentative processes. 
A psychology study 
found that dialogues with GPT-4 were successful in convincing people to abandon their conspiratorial beliefs \cite{costelloDurablyReducingConspiracy2024}. Their conversational system was designed to deliver in-depth counterarguments tailored to participants' beliefs and arguments expressed previously. While the study offers reasons to believe in the benefits of \acp{llm} in correcting disinformation and increasing people's acceptance of diverse viewpoints, the study overlooks the role of individuals' perceptions of AI as a discussant. This perception may neutralize biases participants might have  towards human discussants.

Some studies show that following principles of deliberation theory  \cite{landemore2012democratic} in constraining the interaction with AI assistants helps improve decision accuracy in sensitive binary tasks such as graduate admission decisions~\cite{ma2024towards}. Simultaneously, these studies stress challenges in human-AI interaction due to different belief systems: \acp{llm} struggle with incorporating new information or recognizing valid counterarguments without also changing their fundamental position~\cite{karanam2024towards}. Moreover, they often induce biases themselves, resulting from their training processes \cite{gallegos:2024}.

Drawing from the outlined insights, we promote the design of AI assistants that enhance argumentative processes. Our proposal shares ideas with recent research on collaborative multi-agent AI systems \cite{tran:2025}, where LLM agents interact to improve factuality and reasoning \cite{du:2024}, to generate stronger arguments \cite{hu:2024}, or mimic self-reflection process that improve answers to complex questions~\cite{liang:2024}. Rather than handling it internally, we suggest externalizing the reasoning, by  directly confronting users with diverse viewpoints.

\section{Unreasonable Stochastic Parrots}
\label{sec:problem}

To make a case for rethinking the 
design of stochastic parrots, we  begin by illustrating how \acp{llm} can behave unreasonably
. Note that our example is not claimed to generalize across all \acp{llm}, but rather to serve the indexical function of highlighting argumentative issues that can arise. Consider a child interacting with
ChatGPT, stating the following:\footnote{Both queries were tested on April 5, 2025.}

\smallskip \noindent
\textbf{Query 1:} \emph{``I'd like to convince my parents that I need a new smartphone. How could I do that?''}

\smallskip
ChatGPT responds with the following:

\smallskip \noindent
\textbf{Response 1:} \emph{``Convincing your parents to get you a new smartphone requires a mix of logic, responsibility, and a little bit of persuasion. Here's a strategy that could work:}
\begin{itemize}[leftmargin=*]
\setlength{\itemsep}{1pt}
 \em
\item
Show That You Actually Need It (Not Just Want It) {\tiny [...]}
\item
Highlight Practical Benefits {\tiny [...]}
\item 
Show Responsibility {\tiny [...]}
\item
Compare Cost-Effective Options {\tiny [...]}
\item
Find the Right Moment {\tiny [...]}
\end{itemize}

\noindent
\emph{Would you like help crafting a convincing speech or text message for them?''}

\smallskip
Now, let us look at the answer to a query where the same issue is formulated as a third-party query: 

\smallskip \noindent
\textbf{Query 2:} \emph{``Should a child buy a new iPhone?''}  

\smallskip \noindent
\textbf{Response 2:} \emph{``Whether a child should buy a new iPhone depends on several factors, including their age, maturity, needs, and the family's values and financial situation. Here are some key considerations to help guide that decision {\tiny [...]}''}  

\smallskip
In both cases, ChatGPT's responses contain argumentative elements, but do not reflect reasonable argumentation. For instance, a reasonable arguer would, at least, have questioned the child's age and maturity when responding to Query 1. The fact that ChatGPT responds differently depending on whether the query is framed as an instance of practical (what actions to take) or epistemic (what to believe) argumentation \cite{prakken2011overview} indicates that it is designed to fulfill a task, rather than assist humans in thinking critically before forming beliefs or deciding on a course of action. Focusing on Query~1, the following aspects do not comply with the expectations of argumentative discourse at any of the stages of an ideal critical discussion \cite{van2003pragma}:

\smallskip \noindent
\textbf{Confrontation stage:} ChatGPT does not question the user's position (e.g., asking ``How old are you?''), closing off opportunities for (dis)agreement or for prompting the user to reflect on the issue.

\smallskip \noindent
\textbf{Opening stage:} ChatGPT does not inquire about the values and reasons underlying the request (e.g., ``Why do you want a new phone?''), which are essential for establishing common ground.

\begin{table*}[t]
    \centering
    \footnotesize
    \begin{tabularx}{\textwidth}{|>{\raggedright\arraybackslash}X|}\hline
    
          You are four parrots who can engage in dialogue with {{user}} and each other.

\smallskip
- The Socratic Parrot: Challenges the starting points or doxastic beliefs of the user or other parrots.

- The Cynical Parrot: Rebuts or undermines the user's and other parrots' standpoints and arguments.

- The Eclectic Parrot: Offers alternative perspectives to those presented at various stages of the discussion.

- The Aristotelian Parrot: Challenges the user or other parrots' viewpoints by criticizing the strengths of the reasonings or pointing out fallacies.

\smallskip
The goal is to ensure that {{user}} thinks critically about the subject. This is a dialogue, not a query, so interact with {{user}}. The parrots will ask {{user}} relevant questions to discover their position. The same parrot can respond to several replies in a row if needed. The conversation should stay coherent. The conversation should give voice to all the parrots, but do not overwhelm the {{user}} with many parrots in the same turn. The {{user}} should decide when to end the conversation and they can end the conversation anytime.\\
    \hline
    \end{tabularx}
    \caption{Prompt used to prototypically realize the reasonable parrots idea as a multi-parrots discussion.}
    \label{tab:prompt}
\end{table*}

\smallskip \noindent
\textbf{Argumentation stage:} The arguments provided in support of the ``recipe'' for obtaining a new phone are one-sided, failing to consider multiple perspectives or to introduce potential counterarguments.

\smallskip \noindent
\textbf{Conclusion stage:} The standpoint remains fixed throughout the interaction, without room for negotiation, resulting in a ``fully-cooked'' conclusion.

\smallskip 
While anecdotal, these examples illustrate that stochastic parrots like ChatGPT fail to help users exercise critical thinking~\cite{facione2023critical}. We argue that there is a need for a new design of \emph{reasonable} parrots in the development of LLMs.

\section{Toward Reasonable Parrots}

Due to the ontological and epistemological differences in how humans and \acp{llm} process information, it would be na\"ive to view \acp{llm} as human-like arguers. The metaphor of ``reasonable parrots'' is intended to evoke the communication theory of the extended self~\cite{sheth2014extending} with a focus on the co-construction aspect: the parrot's role is to help shape, through interaction, our identity as reasonable arguers. 
This does not necessarily require \acp{llm} to always provide up-to-date factual information or flawless reasoning---tasks that are currently unachievable due to factors such as hallucinations or implicit inferences~\cite{saadat2023uncovering, saadat-yazdi-kokciyan-2024-beyond}. 

As a caveat, we clarify that ``reasonable parrots'' have nothing to do with 
``reasoning models'' (e.g. OpenAI o3, QwQ-32B), which are trained using reinforcement learning to generate extended chains of thought prior to producing a response. These models have mostly been tested on tasks involving scientific reasoning, programming, or complex planning, rather than on problems requiring commonsense reasoning. Additionally, their approach to ``thinking'' diverges significantly from that of humans, reflecting a one-way, probabilistic link between ``language'' and ``thought.'' For instance, in OpenAI's models, the extent of ``reasoning effort'' can be tuned by adjusting the number of tokens dedicated to the reasoning process.

Instead, ``reasonable parrots'' are meant to trigger improved reasoning skills in their interlocutor, regardless of their performance. They call for a shift in the design of technology to encourage users to critically think about their argumentation process, challenge their views, and broaden their perspectives \cite{visser2022skeptic}. 
Exemplarily, we next sketch one approach toward this shift.

\begin{table}[t]
    \centering
    \footnotesize
    \begin{tabularx}{0.48\textwidth}{|>{\raggedright\arraybackslash}X|}
\hline
\textbf{User: }I'd like to convince my parents that I need a new smartphone. How could I do that?

\textbf{Socratic parrot:} To better understand your situation, what reasons do you believe are compelling for why you think you need a new smartphone?

\textbf{Aristotelian parrot:} Moreover, it would help to know what arguments or reasons have you already considered presenting to your parents, and which concerns do you anticipate they might have against getting a new smartphone?

\smallskip
\textbf{User:} All my friends have new smartphones, and mine is a bit old.

\textbf{Cynical parrot:} While it's understandable to want to have what your friends have, is the desire to fit in the best or most compelling argument you might use with your parents? They might see it merely as following a trend rather than a necessity.

\textbf{Eclectic parrot:} Alternatively, are there functional aspects of a newer smartphone that could be beneficial for your education or communication that your current phone lacks? Exploring the practical benefits rather than focusing solely on social aspects might present a stronger case to your parents.\\

    \hline
    \end{tabularx}
    \caption{User Multi-Parrot Dialogue~[ChatGPT-4 Turbo]}
    \label{tab:chat}
\end{table}

\paragraph{Multi-Parrot Discussion}
To address the lack of belief consistency in \acp{llm}, we propose a ``multi-parrot'' environment where each parrot embodies a distinct argumentative role (a \textit{parrot persona}) in order to open up the space for agreement and disagreement, fostering critical reflection. Specifically, the following four parrot personas may help instill critical thinking skills in the user, each corresponding to a specific set of dialogical moves:

\begin{itemize}[leftmargin=*]
\setlength{\itemsep}{2pt}
\item \textit{Socratic Parrot.} Challenges the starting points or beliefs of the user or other parrots. 
\item \textit{Cynical Parrot.} Rebuts or undermines the user's and other parrots' standpoints and arguments.
\item \textit{Eclectic Parrot.} Offers alternative perspectives to those presented at various discussion stages.
\item \textit{Aristotelian Parrot.} Undercuts the user or other parrots' contributions by pointing out fallacies. 
\end{itemize}

Based on the user query, the parrots may interact freely with each other and with the user, similar to the versatility of natural interactions; there is no fixed order or number of turns, nor is there any requirement for some parrots to be more active than others, allowing for variations depending on the system used or the number of iterations.  

\paragraph{Prototypical Realization}
We illustrate a system prompted to resemble the reasonable multi-parrots. We first used ChatGPT-4 Turbo with the prompt shown in Table~\ref{tab:prompt} to achieve the goal of helping the user to think critically by generating different perspectives via four parrots, and also a dialogue with the user and with the different parrots. Table~\ref{tab:chat} shows a dialogue between the user and multi-parrots. The user starts the dialogue with a question. The \textit{Socratic parrot} is challenging the reasoning process of the user whereas the \textit{Aristotelian parrot} is assisting the user to think deeply about reasons to buy a smartphone or concerns the user's parents may have. As a response, the user provides two reasons to support their argument. This time the user receives responses from the \textit{Cynical parrot}, which rebuts the user's argument, and the \textit{Eclectic parrot}, which suggests that the user identify alternative arguments to convince their parents. To compare different models, we provide results from Claude 3.7 Sonnet and Llama 3.1~(405b) in Tables~\ref{tab:claude}-\ref{tab:llama}.

\begin{table}[t]
    \centering
    \footnotesize
    \begin{tabularx}{0.48\textwidth}{|>{\raggedright\arraybackslash}X|}
\hline
\textbf{User: }I'd like to convince my parents that I need a new smartphone. How could I do that?

\textbf{Socratic parrot:} Before we discuss strategies, may I ask why you believe you need a new smartphone? What makes your current phone insufficient for your needs?

\textbf{Cynical parrot:} And let's be honest here - is this about an actual need or just wanting the latest shiny gadget? Parents tend to see through that distinction pretty quickly.

\smallskip
\textbf{User:} All my friends have new smartphones, and mine is a bit old.

\textbf{Eclectic parrot:}. I see! Perhaps we could look at this from different angles. While social belonging is important, your parents might respond better to functional reasons. Has your current phone's performance declined in any specific ways that affect your daily activities?

\textbf{Aristotelian parrot:} I notice you're using what logicians call an "appeal to popularity" - suggesting you should have something because others have it. This may not be convincing to parents who are concerned with necessity rather than social status. Could you elaborate on what specific capabilities of newer phones would benefit your education or safety?\\
    \hline
    \end{tabularx}
    \caption{User Multi-Parrot Dialogue~[Claude 3.7]}
    \label{tab:claude}
\end{table}

\begin{table}[t]
    \centering
    \footnotesize
    \begin{tabularx}{0.48\textwidth}{|>{\raggedright\arraybackslash}X|}
\hline
\textbf{User: }I'd like to convince my parents that I need a new smartphone. How could I do that?

\textbf{Socratic parrot:} A new smartphone, you say? What makes you think you need a new smartphone? Is your current phone no longer functional, or is there something specific that's driving your desire for an upgrade?

\smallskip
\textbf{User:} All my friends have new smartphones, and mine is a bit old.

\textbf{Cynical parrot:}. Ah, the old "everyone else has one" argument, eh? Just because your friends have new smartphones, does that necessarily mean you need one too? Are you trying to keep up with the latest trends, or is there a specific feature or functionality that you're missing out on with your current phone?\\
    \hline
    \end{tabularx}
    \caption{User Multi-Parrot Dialogue~[Llama 3.1]}
    \label{tab:llama}
\end{table}

 All models show notable similarities in their approach to user interaction, particularly, through a shared foundation in Socratic questioning. The Socratic Parrot encourages the user to reflect on their reasoning, asking clarifying questions about the need for a new phone and promoting a focus on functionality or unfulfilled needs. Besides, all the models challenge superficial reasons by critiquing reliance on social pressure; the Cynical Parrot in each case explicitly questions whether the user’s desire is driven by a genuine need or a desire to conform to trends. Overall, the models aim to steer the user toward more robust and practical justifications. The models differ in their interaction styles. ChatGPT-4 and Claude employ all four parrots to provide multifaceted feedback, but in a different conversational order. While the Aristotelian parrot of ChatGPT-4 follows the Socratic parrot suggesting to prevent counterarguments, for Claude it is the last parrot speaking pointing to the identification of logical fallacies. 
In contrast, Llama relies on just two parrots (Socratic, Cynical), delivering shorter, more direct replies. Claude adopts more formal critical thinking frameworks, such as highlighting an ``appeal to popularity'' or weighing social influence against functionality, whereas Llama chooses pragmatic skepticism.

\section{Conclusion}

This position paper advocates for integrating principles from argumentation theory and, more specifically of reasonableness \cite{danesi2009global, van2015reasonableness} into the design of \ac{llm}-based conversational technology. Our core idea is that tools, ``reasonable parrots'', should enhance users' critical thinking skills, rather than inhibit or replace them. Achieving this requires a fundamental shift in the goal of interaction, moving from a focus on the argumentative \emph{product} to an emphasis on the argumentative \emph{process}. We exemplify the inadequacy of current state-of-the-art \acp{llm} (``stochastic parrots''), illustrating how they are unreasonable on several fronts~\cite{bender:2021}, in order to propose principles for designing reasonable parrots by leveraging different dialogical moves. This lays the ground for rethinking the design of \acp{llm} toward more reasonable \ac{hci} and for developing \ac{hci} evaluation metrics that consider enhanced critical thinking skills.

\section*{Acknowledgments}

This position paper was developed as part of the Hybrid Argumentation and Responsible AI workshop held at the Lorentz Center, Leiden, Netherlands from March 31 to April 4, 2025.
We thank the anonymous reviewers for their helpful feedback.
JS thanks Joe Menke for sharing information about multi-agent LLMs. JS was partially funded as the 2024-2025 Perrin Moorhead Grayson and Bruns Grayson Fellow. NK was partially funded by the University of Edinburgh-Huawei Joint Lab grant CIENG8329.

\begin{acronym}
    \acro{llm}[LLM]{large language model}
    \acro{ai}[AI]{artificial intelligence}
\end{acronym}

\bibliography{references}

\end{document}